\documentclass[11pt,a4paper]{article}
\usepackage[T1]{fontenc}
\usepackage{authblk}
\usepackage[letterpaper,bmargin=1in,tmargin=1in,lmargin=1in,rmargin=1in]{geometry}
\usepackage{listings}
\usepackage{color}
\usepackage{hyperref}
\usepackage[table]{xcolor}
\usepackage{graphicx}
\usepackage{multirow}
\usepackage{subfigure} 
\usepackage{amsmath,amssymb,amsbsy}
\numberwithin{figure}{section}

\usepackage{amsfonts}

\begin{document}
\date{}
\title{Single-trial P300 Classification using PCA with LDA, QDA and Neural Networks}

\author{Nand Sharma \thanks{sharma@math.colostate.edu}}
\affil{Department of Mathematics, Colorado State University, CO, USA}

\maketitle

\begin{abstract}
 
The P300 event-related potential (ERP), evoked in scalp-recorded electroencephalography (EEG) by external stimuli,
has proven to be a reliable response for controlling a BCI. 
The P300 component of an event related potential is thus widely used in brain-computer interfaces 
to translate the subjects' intent by mere thoughts into commands to
control artificial devices. The main challenge in the classification of P300 trials in electroencephalographic (EEG) data is the
low signal-to-noise ratio (SNR) of the P300 response. To overcome the low SNR of individual trials, it is common practice to average together many consecutive trials, which effectively diminishes the
random noise. Unfortunately, when more repeated trials are required for applications such as the
P300 speller, the communication rate is greatly reduced. This has resulted in a need for better methods to improve single-trial classification accuracy of P300 response. In this work, we use Principal Component Analysis (PCA) as a preprocessing method and use Linear Discriminant Analysis (LDA)and neural networks for classification. The results show that a combination of PCA with these methods provided as high as 13\% accuracy gain for single-trial classification while using only 3 to 4 principal components. 
\end{abstract}

\section{Introduction}

Various neurological diseases can disrupt the neuromuscular channels through which the brain communicates with the external world. In certain cases like hemorrhage in the anterior brain stem or degenerative neuromuscular diseases like amyotrophic lateral scleriosis (ALS), the patients suffer from a total motor paralysis \cite{wolpow2}. This results in a condition known as \textit{locked-in syndrome}, wherein the patient is awake and fully aware but cannot communicate with the outside world due to complete paralysis. For such "locked-in" patients, there is a need for an assistive technology that needs no muscular activity whatsoever.

A brain-computer interface (BCI) is a device that uses brain signals to provide a direct, non-muscular communication channel between brain and the outside world \cite{wolpow1,bci1International,bci2International}.
The idea underlying BCIs is to measure electric, magnetic, or other physical manifestations of the brain activity and to translate these into commands for a computer or other devices \cite{donchin1,hoff}.  


For patients with locked-in syndrome,the P300 event-related potential (ERP), evoked in scalp-recorded electroencephalography (EEG) by external stimuli, has proven to be a reliable response for controlling a BCI~\cite{k2}. In this study we present comparison of some classification methods to classify an EEG signal based on the presence of P300 component.

\subsection{BCI and P300}
Types of BCIs can be broadly classified into two categories---those that use an external stimulus, and those that don't \cite{donchin1}. In the first method, the external stimuli cause changes in neurophysiologic signals called event-related potentials (ERPs) \cite{hoff,erp} which are used to identify a user's response to the stimuli presented. In the second method, users generate certain detectable patterns of neurophysiologic signals by concentrating on a specific mental task. For example, imagination of hand movement can be used to modify activity in the motor cortex \cite{hoff}.

For recording the activity of the brain, the electroencephalogram (EEG) is the method of choice for a BCI due to their fast responsivity and covariation with cognitive processes \cite{wolpow2}. Although invasive methods that use electrocorticography (ECoG) signals using implanted electrodes are more accurate, the non-invasive methods are more attractive because of their ease of use by patients; the non-invasive methods have also been shown to be comparable to implanted electrodes ~\cite{k5} when used with appropriate machine learning algorithms.

The EEG non-invasive recordings are done from a set of electrodes placed directly on the scalp 
using the International 10-20 system (Jasper, 1958) \cite{i20}.
In this work, the data is obtained through non-invasive electroencephalographic (EEG) recordings. Moreover, the experiments are run on EEG data using only a subset of 8 electrodes that have already been found to be meaningful for P300 classification~\cite{k5}. A smaller number of electrodes is also better for practical reasons of lower cost and higher usability for target patients. These 8 electrodes used for this work are 'F3','F4','C3','C4','P3','P4','O1','O2'.

\begin{figure}[!htbp]

	\begin{center}
    \includegraphics[width=6in]{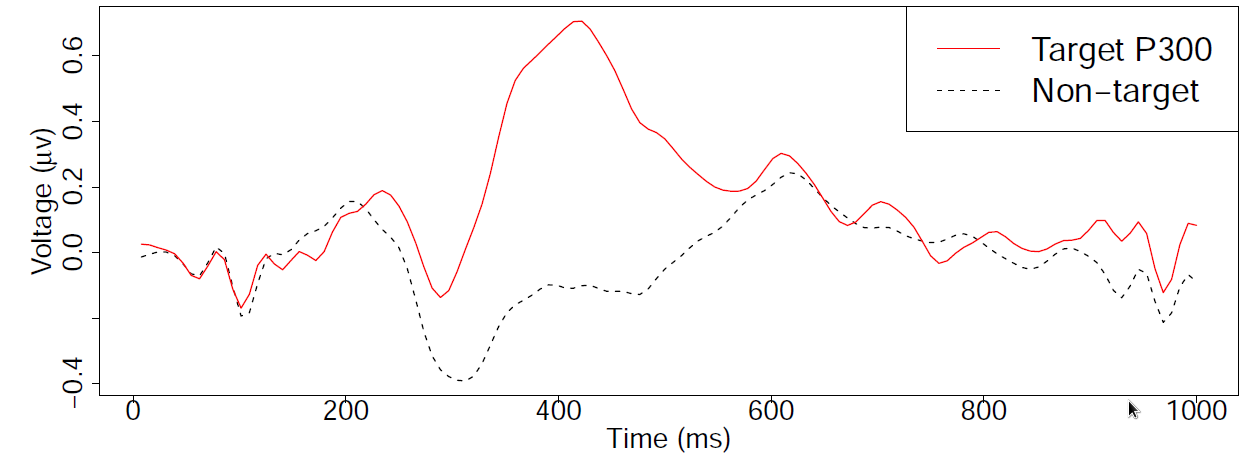}
    \caption{P300 ERP. The image is taken from 				\protect\cite{zach}.}
    \label{fig:p300}
  \end{center}
\end{figure}

A P300 ERP is characterized by a positive peak about 300ms after the stimulus onset \cite{wolpow2,erp} (Figure ~\ref{fig:p300}). It is elicited when subjects encounter a rarely occurring, but expected, stimulus among the presented stimuli. If subjects are assigned the task of assigning a category to each of the stimuli in a series of stimuli of two types, and if one of the two types occurs rarely, a P300 ERP is seen in the EEG \cite{k1} for the rare stimuli. This experimental paradigm based on extensive research has been called 'oddball' paradigm \cite{k1}, the rarely occurring stimuli being the 'oddballs'.

Farwell and Donchin utilized this characteristic of the P300 to design a BCI in 1988 \cite{k1} that is called a 'P300 Speller'. This first BCI of its kind lets a user type one letter at a time using the EEG signals captured through an electrode cap, and needs no muscular movement. There has been a lot of research following their work on improving their setup and many variations of the setup have emerged. As definitions for P300 speller paradigm vary in literature, it is made clear that for this work, each of these windows of EEG will be referred to as a 'subtrial' and a set of row (column) flashings (six in number which contain one target row(column)) will be called a trial \cite{k1}. 

One of the main challenges in P300 classification is the low signal-to-noise ratio (SNR). As the EEG recorded from scalp contains a lot of noise from ongoing electrical activity in the brain, a P300 is hard to separate from the resulting noisy signal. The problem of low SNR is usually overcome by
averaging together many subsequent trials which cancels out most of the noise, and makes P300 detection possible. But this approach of averaging comes at the cost of reduced communication rate. Farwell and Donchin reported in their pioneering work that it needed averaging of between 20 and 40 trials to achieve an accuracy of over 80\% \cite{k1}, with a communication rate of about 12 bits per min or about 2.3 characters per minute. Although their pioneering work did establish the feasibility of a P300 based speller, the communication rate was painfully slow for a practical use. This lead to much work in the last few decades which focused on improving the accuracy and communication speed of a P300-based BCI like P300 speller. So, one of the focus areas of this research has been the development of algorithms that can reduce the number of trials required to achieve a reliable P300-based BCI. But as it is amply demonstrated that averaging of trials stabilizes the P300 amplitude by removing noise, the challenge is also seen as improving single trial accuracy.
\vspace{-0.1in}
\subsection{Related Work}
The last few decades have seen a lot of research on P300 classification, with the eventual aim of achieving single-trial based reliable P300 BCI \cite{donchin1}. But a  literature survey of this research does not show any single P300 classification method to be the state of the art. Krusienski, et al., \cite{wolpaw3} report the results of a comparison of different classifier algorithms, which shows that stepwise linear discriminant analysis (SWLDA) and support vector machines (SVMs) perform well compared to the other classifiers.
In \cite{kru} by using SWLDA as the classification method, Krusienski, et al., achieved at least 60\% accuracy for all participants. Three of the five participants performed above 90\% accuracy with averaging about 15 trials. Sellers and Donchin \cite{k3} achieved comparable average results for healthy and ALS patients using SWLDA. Serby, et al., \cite{k4} used matched-filtering with independent component analysis (ICA) to achieve a communication rate of 5.45 symbols/min with an accuracy of 92.1\% which averaged roughly 15 trials. When the detection was made in real-time by online testing with the same six subjects, the average communication rate achieved was 4.5 symbols/min with an accuracy of 79.5\% which averaged roughly 18 trials. On the same lines, a survey of submissions to BCI Competition II and BCI Competition III shows that several very different approaches like SVM, ICA, LDA, peak-picking methods were able to achieve 100\% accuracy using between 4 and 15 averaged trials \cite{bost,compBci1,compBci2} on the BCI Competition II data. 

This short list shows that there are many different approaches that all
work well, and also that one particular method is not clearly better than others. Then there are other challenges to P300 classification, namely subject-dependence of EEG and even a session-to-session variation in EEG responses of the same subject. A recent review of the BCI field by Mak, et al., \cite{bcireview1} concludes that a lot more work is still needed to create a truly reliable P300 speller. 

In this work, we use Principal Component Analysis (PCA) as a preprocessing method and use Linear Discriminant Analysis (LDA), Quadratic Discriminant Analysis (QDA) and neural networks for classification. PCA has been shown to work well for P300 classification \cite{zach,pca1}. In \cite{zach}, the author compared various blind source separation methods as preprocessing methods for P300 classification, and it was seen that PCA generally worked better than the other methods like Independent Component Analysis (ICA) and Maximum Noise Fraction (MNF) for P300 classification. This work builds upon the work done by Cashero \cite{zach} and tests if classification methods other that Support Vector Machines (SVM) will work well in conjunction with PCA for P300 classification.

Among the classification methods, the choice of LDA is also driven by success of this method and its variants like Stepwise LDA (SWLDA) for P300 classification. Although SWLDA has been shown to work very well for higher accuracy, it is an expensive method with high processing time and is therefore not considered suitable for an online P300 speller system \cite{donchin1}. Considering that, we have preferred to consider LDA which is lightweight and provides good performance generally for P300 classification. QDA is chosen to compare LDA with, and to see whether a linear or a nonlinear method works better for P300 classification. Our choice of Neural Networks (NN) as a method of study for P300 classification is primarily driven by our hypothesis that they should be a good choice for P300 classification. There have been studies employing Neural Networks for P300 classification \cite{pca1, nn2}, and many have shown promise. There is very little work done using PCA and NN as a combination for P300 classification. As NN provide a lot of flexibility in terms of how the 'derived' data is created based on flexibility in the number of hidden units, it could be a useful method for noisy data like EEG, especially if PCA provides a good set of source components. SVM is another method which has been shown to work well for P300 classification. We don't include that in this study due to the fact that a study using PCA with SVM has already been covered in \cite{zach}. To keep our focus on the effect of PCA on these classification methods, we have tested only single trial accuracies. 

The paper is laid out as follows. Section 2 develops the mathematical background for all the algorithms that are used in the experiments
for this work. The classification methods, LDA, QDA, Neural Networks, and the optimization algorithm used for NN and PCA are developed in this section. Section 3 describes the datasets used and details the steps used in the experiments. Section 4 explains the experiments as they were performed and what was learned along the way, and the results of the experiments. Section 5 concludes by providing a summary of the findings and identifies avenues for future work.

\section{Classification and Preprocessing Methods}

\subsection{LDA and QDA }

LDA and QDA \cite{trevor2001elements, bishop} belong to a class of classification methods that model \textit{discriminant functions} for each class and then classify a given data sample to the class with the largest value for its discriminant function. LDA and QDA  model the posterior probabilities for the purpose of defining these discriminant functions. Representing a data sample by variable $X$ and the class label by variable $C$, $X \in \mathbb{R}^{d}$ where $d$ is the dimension of data sample $X$ or equivalently represents the number of features or predictors in each sample $X$, and $C \in \lbrace1,2,...,K\rbrace$ where there are $K$ classes. We thus need the class posteriors $p(C\mid X)$ for a given $X$ and $C$. Suppose $f_k(x)$ is the class-conditional density of $X$ in class $C = k$, and let $\pi_k$ be the prior probability of class $k$, with $\sum_{k=1}^K \pi_k = 1$. Applying the Bayes theorem gives us 

\begin{equation}\label{eq1}
p(C= k|X = x) = \frac{f_k(x)\pi_k}{\sum_{l=1}^K f_l(x) \pi_l}
\end{equation}
 
While many different models can be used to model the class-conditional density, LDA and QDA use multivariate Gaussians to model each class density so that 

\begin{equation}\label{eq2}
f_k(x) = \frac{1}{(2\pi)^{d/2} |{\Sigma_{k}}|^{1/2}}e^{-1/2(x-\mu_k)^T\Sigma_{k}^{-1}(x-\mu_k)}
\end{equation}
where $\Sigma_k$ and $\mu_k $ are Covariance matrix and the mean of the 
Gaussian distribution.
If we assume that the classes have a common covariance matrix given by 
$\Sigma = \sum_{k=1}^K \frac{N_k}{N} \Sigma_k$, where $N_k$ is the class k samples and $N$ is the total number of samples of all classes, then we get LDA. In comparing two classes $k$ and $l$, it is sufficient to look at the log-ratio

\begin{equation}\label{eq3}
\ln \frac{p(C = k\mid X = x)}{p(C = l\mid X = x)} = \ln \frac{f_k(x)}{f_l(x)} + \ln \frac{\pi_k}{\pi_l}
\end{equation}
Plugging in the $f_k(x)$ and $f_l(x)$ as defined in \eqref{eq2}, we get

\begin{equation}\label{eq4}
\ln \frac{p(C = k\mid X = x)}{p(C = l\mid X = x)} = \ln \frac{\pi_k}{\pi_l} - \frac{1}{2}(\mu_k - \mu_l)^T \Sigma^{-1} (\mu_k - \mu_l) + x^T\Sigma^{-1}(\mu_k - \mu_l) 
\end{equation}
which is an equation linear in $x$. So, equation \eqref{eq4} gives us a decision rule for LDA---if the log ratio is positive, the sample $x$ is classified as belonging to class $k$ and to class $l$ if it is negative. The value being zero implies the decision boundary which is linear in $x$. It's obvious that the \textit{linear discriminant functions} 

\begin{equation}\label{eq5}
\delta_k(x) = x^T \Sigma^{-1} \mu_k - \frac{1}{2}\mu_k^T \Sigma^{-1} \mu_k + \ln \pi_k
\end{equation}
are an equivalent description of the decision rule for LDA. So, the class of a new sample $x$ is simply
\begin{align*}
C(x) = \operatorname{arg\,max}_k \delta_k(x)
\end{align*}

Now if we drop the assumption of a common covariance matrix for all classes, then the convenient cancellation of terms leading to \eqref{eq4} doesn't happen and we get the quadratic discriminant functions (QDA), 

\begin{equation}\label{eq6}
 \delta_k(x) = -\frac{1}{2} \ln |\Sigma_k| -\frac{1}{2}(x-\mu_k)^T
	\Sigma_k^{-1} (x-\mu_k) + \ln \pi_k
\end{equation}

The decision boundary in this case between each pair of classes $k$ and $l$ is described by a quadratic equation given by $\delta_k(x) = \delta_l(x)$.
Again, the class of a new sample $x$ is obtained as 
\begin{align*}
C(x) = \operatorname{arg\,max}_k \delta_k(x)
\end{align*}

The LDA and QDA are also called Generative models, as they make the assumption of Gaussian distribution of the data, and base the classification of samples on that assumption. The parameters of the Gaussian distributions are estimated using the training data. Class priors are calculated based on the number of samples of each class present in the training data, as  
\begin{align*}
\pi_k = \frac{N_k}{N},
\end{align*} where $N_k$ is the class $k$ samples and $N$ is the total number of samples of all classes. The class means are estimated as the means of the data of a particular class in the training data, so
\begin{align*}
\mu_k = \sum_{c_i=k} x_i/N_k
 \end{align*}

where $c_i$ is the class of sample $x_i$.
Similarly $\Sigma_k$ is the covariance matrix for each class based on the data of that particular class in the training data.  

\subsection{Neural Networks}
Neural Networks (NN) \cite{trevor2001elements, bishop} are primarily employed in machine learning as nonlinear regression and classification method. While there are many variations and flavors of NN like Recurrent NN, multi-layer perceptron, etc., \cite{trevor2001elements, bishop, nn1}, we use a single hidden layer NN for this work. This basic neural net, sometimes called the single hidden layer back-propagation network, or two layer perceptron is a two-stage regression or classification model. It consists of an input layer, a hidden layer and an output layer.

In the case of $K$-class classification, there are $K$ output units, and each of the $K$ output units models the probability of class k so that 
\begin{align*}
Y_k \in [0,1] \hspace{0.2cm} \forall \hspace{0.2cm} Y_k, k = 1, . . . ,K 
\end{align*}

Derived features $Z_m$ are created from linear combinations of the inputs, followed by a nonlinear activation function. The output $Y_k$ is modeled as a function of linear combinations of
the $Z_m$,

\begin{eqnarray}
Z_m = \sigma(\alpha_{0m} + \alpha_m^TX),\hspace{0.2cm} m = 1,...,M,\\
Y_k = \beta_{0k} + \beta_k^TZ,\hspace{0.2cm} k = 1,..., K,\\
f_k(X) = g_k(Y),\hspace{0.2cm} k = 1,...,K
\end{eqnarray}\label{eq8}
where $Z = (Z_1,Z_2,...,Z_M)$, and $Y = (Y_1, Y_2,..., Y_K)$.
$\sigma(v)$ is the activation function and is chosen to be a \textit{sigmoid} defined as 
\begin{align*}
\sigma(v) = \frac{1}{1+e^{-v}}
\end{align*} 

The output function $g_k(Y)$ allows a final transformation of the vector of
outputs Y. We choose the softmax function for this :
\begin{equation}
g_k(Y) = \frac{e^{Y_k}}{\sum_{l=1}^K e^{Y_l}}
\end{equation}
which results in a multilogit model, and produces positive estimates that add to one. Treating these outcomes as probabilities for the corresponding class, we use negative log-likelihood as the objective function to minimize. This negative log-likelihood objective function is defined as 

\begin{equation}
LL(\theta) = - \sum_{i=1}^N \sum_{k=1}^K T_{ik} log f_k(x_i)
\end{equation}
where $\theta$ denotes the complete set of weights of the network, which consist of $\lbrace \alpha_{0m} ,\alpha_m,\hspace{0.2cm} m = 1,...,M\rbrace$ and $\lbrace \beta_{0k} ,\beta_k,\hspace{0.2cm} k = 1,...,K\rbrace$, and $T_{ik}$ is the $k^{th}$ component of the target indicator variable $T_i$ for $i^{th}$ training sample $x_i$. Finally, the corresponding classifier is defined as 
\begin{align*}
C(x) = \operatorname{arg\,max}_k f_k(x)
\end{align*}

With the sigmoid activation function and log-likelihood  error function, the neural network
model works as a linear logistic regression model in the hidden units, and all the parameters are estimated by maximum likelihood. But by using the nonlinear transformation $\sigma$, it becomes a non-linear model of inputs $X$. Another interesting aspect of this model is that the number of hidden units can be varied to adjust the non-linearity of the model. If there are no hidden units, the model becomes a simple linear logistic regression model over input data. In this work we use both linear and non-linear versions of the neural network model and call them LR and NLR respectively. 

The error function $LL(\theta)$ can be minimized by variety of approaches. One of the standard approaches is the gradient descent, called \textit{back-propagation} in the neural network setting. But a conjugate gradient method called Scaled Conjugate Gradient \cite{Moller93ascaled} has been more successful and faster for this optimization problem and we use that in this work and we use SCG for this work. 

\subsection{PCA}

Principal Component Analysis, or PCA, \cite{bishop,strang,kirby} (also known as \textit{Karhunen-Loeve} transform) is a technique that is widely used in pattern recognition and machine learning for dimensionality reduction and feature extraction. There are two commonly used derivations of PCA---one that maximizes variance of data, and the one that minimizes the projection error. We develop the maximum variance formulation.

Given a set of data samples $\lbrace x_n, n = 1,...,N \rbrace, x_n \in \mathbb{R}^{D}$, the goal of PCA is to project the data onto a space with   dimensionality $M \leq D$, while maximizing the variance of the projected data. If $v_1$ is the first direction of projection, the variance of the data projected on $v_1$ is given by

\begin{equation}\label{variance}
\frac{1}{N}\sum_{n=1}^N \lbrace v_1^Tx_n - v_1^T\bar{x} \rbrace^2 = v_1^TSv_1
\end{equation}
where $\bar{x} = \frac{1}{N}\sum_{n=1}^N (x_n)$ is the sample set mean, and S is the data covariance matrix defined by 

\begin{equation}
S = \frac{1}{N}\sum_{n=1}^N (x_n - \bar{x})(x_n - \bar{x})^T = v_1^TSv_1
\end{equation}

So, the optimization problem becomes that of maximizing \ref{variance}, or equivalently

\begin{equation}
\max_{v_1} \; v_1^TX^TXv_1 ,\; v_1^Tv_1 = 1
\end{equation} 
which is a constrained optimization problem. Using a Lagrange multiplier $\lambda$, we get

\begin{equation}\label{lagr}
\max_{v_1,\lambda} \; v_1^TX^TXv_1 + \lambda(1- v_1^Tv_1)
\end{equation}
By setting the derivative of \ref{lagr} with respect to $v_1$ equal to zero, we get
\begin{equation}
2Xv_1 - 2\lambda v_1 = 0
\end{equation} 
\begin{equation}\label{lagrSol}
X^TXv_1 = \lambda v_1 
\end{equation}
which means that $v_1$ must be an eigenvector of $X^TX$ with eigenvalue $\lambda$, which also turns out to be a measure of the variance. So, $v_1$ turns out to be a direction of projection that results in maximum variance in the projected data, and is called the first 'Principal Component'. The subsequent directions can be found inductively as follows. Given that we have up to $v_k$ Principal Components, the next Principal Component $v$ can be found by solving the following constrained optimization problem :

\begin{equation}\label{mainlagr}
\max_{v, v \bot v_1,..,v_k}  v^TX^TXv , \; v^Tv = 1
\end{equation}
And this problem can be solved by creating the Lagrangian. Solving Equation \ref{mainlagr} for all eigenvectors is equivalent to computing the singular value decomposition (SVD) of $X$ \cite{strang,kirby,golub}:

\begin{equation}\label{svd}
X = U \Sigma V^T
\end{equation}
where $U \in \mathbb{R}^{N \times N}$, $V \in \mathbb{R}^{D \times D}$ and $\Sigma \in \mathbb{R}^{N \times D}$. In this decomposition, the columns of V, the right singular vectors, are the eigenvectors of $X^TX$ \cite{strang,kirby,golub} which provide us the required 'Principal Components'. Additionally, these column vectors are ordered by the variance they produce when data is projected onto them so that the first column of $V$ is the first Principal Component, the second one the second and so on. In this work, we thus use SVD to derive the Principal Components for our experiments.
	
%
%

\section{Data Acquisition and Representation}

This Section describes the datasets used in this study, as also the methods and EEG recording equipment used. As data representation is an important part of any signal processing method, we also describe the data representation used in this study. A novel way of using channel-subtrials is also defined and explained.
\subsection{Datasets }
 
There are four subjects in the study. The EEG data for subjects 1 and 2 was recorded by BCI laboratory at Computer Science department at Colorado State University \cite{bciCSU}. The data were recorded using the g.Tec g.GAMMAsys system \cite{gtec} with a 8-electrode cap with electrodes located at Fz, Cz, Pz, Oz, P3, P4, O1, O2. This subset of electrodes has been found to be meaningful for P300 classification ~\cite{k5,tobci}. Such small subsets of electrodes are also easier to use for a practical and easy-to-use BCI. Subject 2 was an able-bodied participant in the study, and subject 1 was a subject with C4 complete Spinal cord injury. C4 is a level of Cervical (neck) injury that results in significant loss of function at the biceps and shoulders. While the data for subject 2 was collected in the laboratory, the data for subject 1 was recorded at home.

Three sessions of data collection were performed with both subjects 1 and 2. The subjects count one of three target letters ('b', 'd', 'p') during a session as various other non-target letters are randomly flashed on a screen. This data collection is done as per 'odd-ball' paradigm and the probability of occurrence for the target letter in each session was 0.25. The data were sampled at 256 Hz, with an inter-stimulus-interval (ISI) of 1 second. One session consisted in recording 20 target and 60 non-target subtrials of 1000ms each of EEG data at 8 electrodes.

Data for subjects 3 and 4 is taken from BCI competition III (dataset II) \cite{bcidata}. These experiments being based on the P300 speller as proposed by Farwell and Donchin \cite{k1}, the subjects were presented with a 6 by 6 matrix of characters. The subject's task was to focus attention on characters in a word that was prescribed by the investigator (i.e., one character at a time). All rows and columns are successively and randomly intensified at a rate of 5.7Hz. The objective in this contest was to predict the correct character in each of the provided character selection 'epochs' that consisted of 15 sequences---each sequence consisting of 12 flashings---6 rows and 6 columns. The data was collected and bandpass filtered from 0.1-60Hz and digitized at 240Hz. After intensification of a row/column, the matrix was blank for 75ms. Row/column intensifications were block randomized in blocks of 12. The sets of 12 intensifications were repeated 15 times for each character epoch (i.e., any specific row/column was intensified 15 times and thus there were 180 total intensifications for each character epoch). Each character epoch was followed by a 2.5 s period, and during this time the matrix was blank. While data for subjects 3 and 4 is recorded at 64 electrode locations, this work uses data only from 8 locations which is the same set of electrodes used for subjects 1 and 2---Fz, Cz, Pz, Oz, P3, P4, O1, O2.
  	
\subsection{Data Processing and Representation}
\begin{figure}[ht]
\begin{minipage}[b]{0.45\linewidth}
\centering
\includegraphics[width=\textwidth]{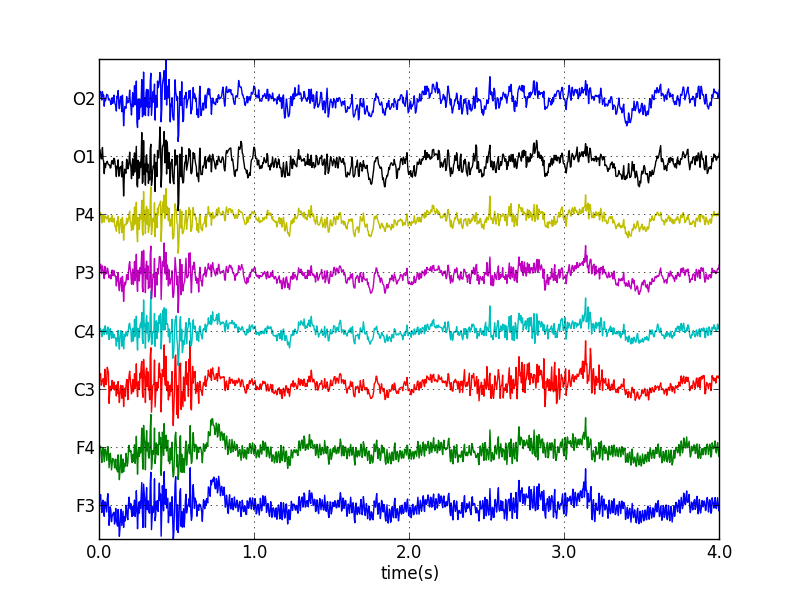}
\caption{4s Original Data sub1.}
\label{fig:OriginalData_s11_b}
\end{minipage}
\hspace{0.5cm}
\begin{minipage}[b]{0.45\linewidth}
\centering
\includegraphics[width=\textwidth]{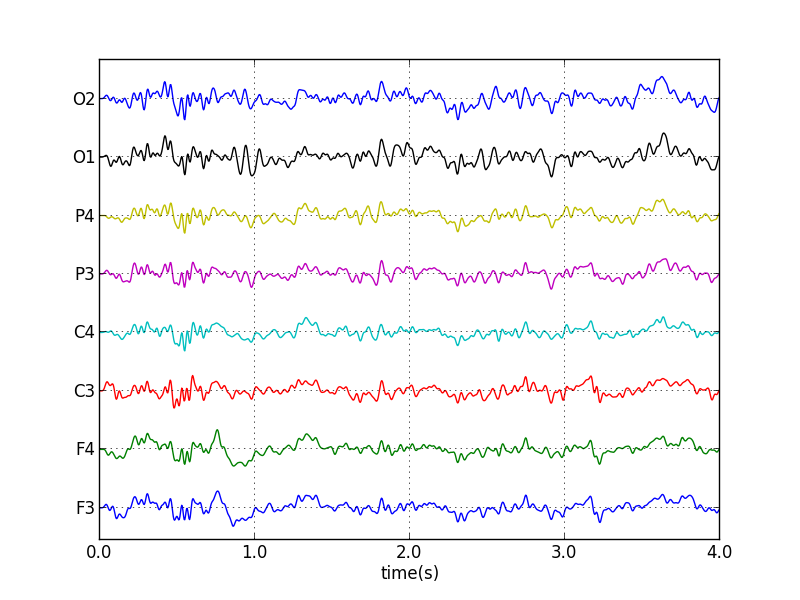}
\caption{4s Data sub1 after bandpassing.}
\label{fig:OriginalData_s11_c}
\end{minipage}
\end{figure}

The original data for all the four subjects comes as a continuous EEG for an entire recording. 
All the data was then bandpass-filtered from 0.23 Hz to 30 Hz. The data were then normalized for zero mean and unit variance. Figures \ref{fig:OriginalData_s11_b} and \ref{fig:OriginalData_s11_c} show a 4 second window of data before and after the bandpass filtering. The bandpass filtering was done using Butterworth bandpass filter \cite{butter}. 
This data is then sliced to separate the target and non-target subtrials for each channel. Each Dataset was thus reshaped into a matrix with each row representing a channel-subtrial- 256 datapoints as a time series for subjects 1 (sub1) and 2 (sub2), and 240 for subjects 3(sub3) and 4(sub4). The channel-subtrials used for classification therefore consist of one-second long windows after each stimulus onset that
are extracted from the continuous signal in each data segment.

Our approach for data representation involves considering a time series from each of the eight channels as a separate subtrial for the purpose of initial classification. What this means is that a subtrial that in the usual sense consists of eight different time series, one for each channel, is split into eight different subtrials in our data matrix. This is done with the view that P300 response is present in each of the eight channels considered  although it varies in degree and has some phase difference, and that our preprocessing method should be able to get common sources to these responses at different channels. We call these channel subtrials. Once the training of the algorithm is done, and results for the test set are collected for each channel sub trial, these results are aggregated to arrive at a classification for the 'overall' subtrial. This aggregation is done using a voting method. 

The P300 data is generally unbalanced in positive and negative examples, as every trial contains many more negative examples than the positive ones in accordance with the 'odd-ball' paradigm. This usually tends to bias a classifier in favor of the negative examples.  So, an equal number of subtrials of target and non-target EEG were used for these experiments (20 for sub1 and sub2, and 30 for sub3 and sub4).

\section{Experiments and Results}
	
Experiments were designed to test and compare the various approaches selected for this work with and without PCA. So, the following workflow was followed.

\begin{enumerate}
\item Classification performance of the four classifiers---LDA, QDA, LR and NLR---was recorded on the raw data, i.e., the data that has not been transformed with PCA. In this, no feature selection is done, so that all the original features are used.
\item Then classification performance is tested after performing PCA on the data. There are two procedures in this set of experiments---one with forward selection of principal components across all the components, and one where forward selection is done on selected number of top components as per the magnitude of the singular values.
\end{enumerate}
\subsection{Without PCA}

We start experiments by testing our four methods on the raw data, i.e., the data without PCA transformation for the four subjects. This would serve as a benchmark for the experiments to follow using the PCA. For these experiments and that follow, the following common approach was followed.

\begin{enumerate}
\item The datasets were randomly partitioned into training and test sets in the ratio of 80:20. As we consider channel-subtrials as separate subtrials during training and classification before voting, the data was partitioned to ensure we place complete set of 8 channel-subtrials belonging to a subtrial together either in the training or test sets.
\item The algorithms are trained on the data comprising the channel-subtrials, and tested on the test channel-subtrials.
\item The accuracies on the channel-subtrials from the 8 channels are aggregated together using the voting method to decide on the final class of the actual subtrial.
\item The above process is repeated 20 times and the accuracies over the 20 runs are averaged to obtain the final accuracies.
\end{enumerate}

For NLR, the training set was partitioned to create a validation set. The accuracies on the validation set were used to choose the number of hidden units. In the pilot studies on all four subjects, a range of number of hidden units from 2 to the number of features was tested on the validation set. It was observed that the validation accuracies generally were the best at number of hidden units equal to the number of features in the dataset. So, throughout the experiments, number of hidden units for a dataset have been taken as number of features used for classification. 

The accuracies obtained on the raw data for the four subjects is shown in Table \ref{table: rawAcc}. First thing to note is that QDA did not work for any of the subjects, the problem being that the covariance matrices became singular, that gave a runtime error. So, the accuracies for QDA have been left blank for raw data. The problem of sample covariance matrices being singular occurs when sample size is less than the number of features, which happens to be the case when using QDA on raw data. The same problem was not usually encountered in the case of LDA as the averaging of the sample covariance matrices of the two classes resulted in the average covariance matrix being non-singular. In the subsequent experiments, a small subset of features was always used, which ensured that this problem is not encountered.

Analyzing the results, we observe as generally expected in P300 classification that no single method is the best across all 4 subjects. LR works the best for subjects sub1 and sub4, but NLR also works the best for two subjects sub3 and sub4. It's only for sub2 that neither of the NN versions, linear and nonlinear, work well. And it is LDA that works the best for sub2. So, across subjects, we can not generalize if linear or nonlinear methods are a clear winner. But for sub1, the performance of the linear version of NN, the LR, is far superior than the other methods. The best accuracies obtained for all the subjects are near 50\% which is the expected accuracy for a random classifier for single trials \cite{k4,zach} except sub1 which gets a very high accuracy for single trials.  

\begin{table}[ht]
\caption{Accuracies with all four methods (without PCA)}
\label{table: rawAcc}
\centering
\begin{tabular}{c c c c c c}
\hline\hline
\vspace{4 mm}
Subjects $\longrightarrow$ & sub1 & sub2 & sub3 & sub4\\ [0.5ex] 
Algorithms $\downarrow$ \\

\hline 
LDA & 47.33 & \textbf{51.25} & 49.58 & 44.58  \\
QDA & - & - & - & - \\
LR & \textbf{65.62} & 48.25 & 47.08 & \textbf{51.25} \\
NLR & 58.75 & 42.12 & \textbf{50.83} & \textbf{51.25}  \\
[1ex]
\hline\hline
\end{tabular}
\vspace{4 mm}

\end{table}
\vspace{1 cm}
		
\subsection{With PCA}

\begin{figure}[ht]
\begin{center}
\includegraphics[width=7in]{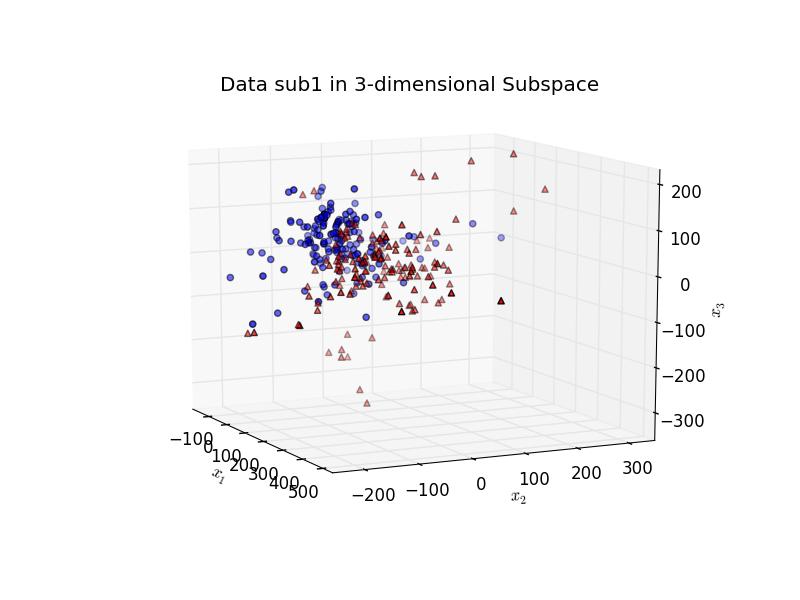}
\caption{sub1 data projected on 3-D subspace.}
\label{fig:3d_s11_a}
\end{center}
\end{figure}

The next step in our experiments was to use PCA for feature extraction. We start our experiments with PCA by getting the Principal Components (PCs) of the training data using SVD. Then we plot the projection of training data onto first 20 components for each of the four datasets. Looking at these projections, we choose the top 3 PCs through visual inspection---for sub1 these turned out to PCs 2,3, and 4. Projection of sub1 data onto a 3-dimensional subspace of these 3 PCs is shown in Figure \ref{fig:3d_s11_a}.

After a similar analysis for subjects sub2, sub3 and sub4, we choose the top 3 components for each of them. Then using the chosen three components, we run our classification experiments. The results of these experiments are shown in Table \ref{table: PCArankAcc3}.

\begin{table}[ht]
\caption{Accuracies for PCA-transformed data using only 3 visually chosen PCs}
\label{table: PCArankAcc3}
\centering
\begin{tabular}{c c c c c c}
\hline\hline
\vspace{4 mm}
Subjects $\longrightarrow$ & sub1 & sub2 & sub3 & sub4\\ [0.5ex] 
Algorithms $\downarrow$ \\

\hline 
PCA+LDA  & \textbf{55.0} & 55.0 & \textbf{49.16} & \textbf{55.00}  \\
PCA+QDA & 52.5 & 46.0 & \textbf{49.16} & 51.6 \\
PCA+LR & 41.6 & 40.8 & 45.0 & 50.83 \\
PCA+NLR & 41.6 & \textbf{56.6} & \textbf{49.16} & 53.33  \\
[1ex]
\hline\hline
\end{tabular}
\vspace{4 mm}

\end{table}	

While this method improved the best accuracy for sub2 and sub4, for sub1 and sub3 the accuracies decreased with almost all the methods. Although the accuracy gains for sub2 and sub4 are impressive, this method is neither scientific nor desirable for a BCI because of its reliance on human intervention and visual judgment. Moreover, there is a limit to the number of PCs that we can inspect visually. But, it does show that if we are able to pick the right components, we could get accuracy gains for some subjects.    

To make the process of choosing the PCs thorough and to automate the process, we use the method of 'forward selection' (FS) of the PCs in each of the four algorithms. The forward selection is done using 3-fold cross-validation to choose the PCs that give the best accuracy for the validation set. A new PC is thus added to the set of PCs at each iteration. Finally, the minimal set that gives the best accuracy on the validation set is chosen. As this method is very expensive, we pick and test validation accuracies till only the top 50 components are chosen. This also ensures that data sample size being less than the dimensionality doesn't give singular covariance matrices in the case of QDA. This method thus iterates through all the PCs at each iteration and picks the best PC at each iteration. The results of this method are shown in table \ref{table: PCAfsAcc}.

\begin{table}[ht]
\caption{Accuracies for PCA-transformed data with number of PCs chosen using forward selection(Number of Principal Components in braces)}
\label{table: PCAfsAcc}
\centering
\begin{tabular}{c c c c c c}
\hline\hline
\vspace{4 mm}
Subjects $\longrightarrow$ & sub1 & sub2 & sub3 & sub4\\ [0.5ex] 
Algorithms $\downarrow$ \\

\hline 
PCA+LDA  & 54.37(11) & 53.12(14) & 49.58(16) & 48.75(14)  \\
PCA+QDA & 49.37(18) & \textbf{55.12}(18) & 50.83(15) & 47.91(12) \\
PCA+LR & 42.50(6) & 50.46(5) & 47.08(3) & 50.00(5) \\
PCA+NLR & \textbf{55.00}(4) & 48.03(5) & \textbf{51.25}(20) & \textbf{50.83}(5)  \\
[1ex]
\hline\hline
\end{tabular}
\vspace{4 mm}

\end{table}		

The results using forward selection of Principal components are not found to be as encouraging as expected. The challenge lies in selecting the number of top components based on the validation accuracies. The method chooses the number of components that give the best validation accuracy, but this doesn't necessarily work for the test data. The reason for that is the extreme noise in the data due to which the validation set accuracies don't generalize well to test data. The FS would pick such components that might be modeling the noise of the validation set which obviously will throw the classifier off on the test data. Then there is also over-fitting of training data in the case of non-linear methods as can be seen for NLR. In addition to these problems, there is the matter of complexity and runtime. The FS is a very expensive method as each iteration of choosing a new component has to iterate through all the remaining set of components. 

Considering the above pitfalls of the FS method, we need to try a different method for selecting a good subset of PCs. Looking back at the results of using top 3 PCs chosen visually from top 20 components, we try modifying the forward selection algorithm to select the components only from a set of an empirically chosen number, say $n$, of top PCs based on the magnitude of their singular values. This amounts to just considering the top $n$ PCs as they are ordered already by magnitude of their singular values. As visually chosen 3 components had shown good results for 2 out of 4 subjects, we try $n = \lbrace5,10,15,20\rbrace$. After looking at the results thus obtained, it was observed that using $n=5$ works the best. The results of using FS on a restricted set of top 5 PCs is shown in table \ref{table: PCArankAcc5}.

\begin{table}[ht]
\caption{Accuracies for PCA-transformed data choosing the best among only top 5 PCs (Number of Principal Components in braces)}
\label{table: PCArankAcc5}
\centering
\begin{tabular}{c c c c c c}
\hline\hline
\vspace{4 mm}
Subjects $\longrightarrow$ & sub1 & sub2 & sub3 & sub4\\ [0.5ex] 
Algorithms $\downarrow$ \\

\hline 
PCA+LDA  & 55.0(2) & 48.0(3) & 45.0(2) & 47.5(3)  \\
PCA+QDA & 55.0(2) & 52.7(3) & 40.83(2) & \textbf{53.3}(1) \\
PCA+LR & 41.66(3) & \textbf{57.5}(2) & 56.9(4) & 46.66(2) \\
PCA+NLR & \textbf{60.0(3)} & 40.0(3) & \textbf{63.05}(4) & 50.83(4)  \\
[1ex]
\hline\hline
\end{tabular}
\vspace{4 mm}

\end{table}	

The results obtained using FS on the restricted set of top 5 PCs show an improvement in accuracy for all the four subjects when compared with regular FS. The method gives overall best accuracies for sub2 and sub3. For sub4 also, the accuracy is better than all methods except visual selection of PCs which can be discounted as that's not a practical method for an online BCI. So, discounting the visual selection method, we get best accuracies for three subjects out of four using this method of restricted FS. In addition, this method comes with greatly reduced data dimensionality. While FS selected as high as 18 PCs, this method needed no more than 4 components to achieve much better results.  

\section{Conclusions }

\subsection{Summary of Results }
As the experiments have shown and as is supported by the literature, the success of a method for P300 classification depends a lot on the subjects. While LR gave exceedingly good results for sub1 without PCA, sub3 had the best accuracy of all the methods with NLR + PCA (with restricted FS). NLR + PCA (with restricted FS) also gave the second best accuracy for sub1 at 60\%. 

For sub2, on the other hand, it was the linear version of NN---the LR (with restricted FS)---that gave the best accuracy. For sub4 alone, the visually chosen 3 PCs gave the best accuracy. But overall it's clear that PCA did help in improving the accuracy of classification of single trials in all subjects but sub1. The best results except in the case of sub1 all came with a much reduced dimensionality of data---the dimensionality was a maximum of 4 for all these subjects.	So, PCA feature selection not only increased the classification accuracy but also reduced the execution time of the algorithms by the resulting dimensionality reduction. 

Among the wider distinction between linear and nonlinear methods, the results were split. While sub1 and sub2 got the best accuracies using linear LR, sub3 and sub4 got the best results with QDA and NLR---both nonlinear methods. The reason for this seems to be different data acquisition methods used for the two datasets. While sub1 and sub2 data was collected with an ISI of 1000ms, data for sub3 and sub4 was collected with an ISI of just 175ms. The short ISI resulted in the overlapping of P300 amplitude of the EEG which happens about 300ms with the new stimulus induced signal, which also explains why the grand average signal for sub3 and sub4 was not as similar to a P300 as it was for sub1 and sub2. The overlapping of signals from many stimuli seems like the reason that makes these two datasets more difficult to be linearly separable. Another observation is that the method of treating each channel-subtrial as a separate subtrial for the purpose of classification works well for P300. The hypothesis of PCA being able to extract the relevant source components from this channel-subtrial dataset is also validated by the results. Also for sub1 and sub2, the PCA captures the variance across channels into a component (it was the first PC for sub1). Then the classifier would ignore that component as it won't give it any significant discriminatory value for the purpose of classification. Such variance was not that pronounced in the case of sub3 and sub4. 
	
\subsection{Future Work }
This work has many avenues for future work. As discussed earlier, as the P300 classification depends a lot on the subjects, it would be useful to run these same experiments on data from more subjects to see how well the results generalize across subjects.
Then for the same subjects other classification methods especially variants of LDA, such as Fisher's linear discriminant (FLD), stepwise linear discriminant analysis (SWLDA), and regularized Discriminant Analysis, can be tried to see if the results could be further improved. Another interesting thing to try would be to compare our method of channel-subtrial based data with other methods of data representation in spatio-temporal and frequency domains. Also, the methods used in this work can be studied in more depth. For example, it's well known that LDA and QDA require a certain minimum number of data samples for optimal performance. It would be interesting to see if and by how much could the performance further improve if more number of samples are used for training the classifiers used in this work. For NLR, we could also try multiple hidden layers to see how that would work for P300 classification. 
\clearpage

\bibliographystyle{plain}
\bibliography{thesisNand}



%
%


%
%


\end{document}